\documentclass{article}

\usepackage{microtype}
\usepackage{graphicx}
\usepackage{subfigure}
\usepackage{booktabs}
\usepackage{amsmath}
\usepackage{dsfont}
\usepackage{hyperref}
\usepackage{xurl}           
\usepackage{pgfplots}
\pgfplotsset{compat=1.16}
\usepackage{tikz}
\usepackage{xcolor}
\usepackage{multirow}
\usepackage{algorithm}      
\usepackage{algorithmic}    
\usepackage{listings}       

\usepackage[accepted]{icml2019}

\definecolor{colBlue}{RGB}{31,119,180}
\definecolor{colOrange}{RGB}{255,127,14}
\definecolor{colGreen}{RGB}{44,160,44}
\definecolor{colRed}{RGB}{214,39,40}
\definecolor{colPurple}{RGB}{148,103,189}

\definecolor{revcolor}{RGB}{0,82,155}    
\definecolor{notecolor}{RGB}{200,30,30}  

\newif\ifshowrev
\showrevfalse   

\ifshowrev
  \newcommand{\rev}[1]{\textcolor{revcolor}{#1}}
  \newcommand{\note}[1]{\textcolor{notecolor}{\textbf{[NOTE: #1]}}}
  
  \newenvironment{revblock}{\color{revcolor}}{\ignorespacesafterend}
\else
  \newcommand{\rev}[1]{#1}
  \newcommand{\note}[1]{}
  
  \newenvironment{revblock}{}{}
\fi

\lstdefinestyle{prompt}{
  basicstyle=\ttfamily\scriptsize,
  breaklines=true,
  breakatwhitespace=false,
  columns=fullflexible,
  keepspaces=true,
  frame=single,
  framerule=0.3pt,
  rulecolor=\color{gray!55},
  backgroundcolor=\color{gray!6},
  xleftmargin=3pt,
  xrightmargin=3pt,
  aboveskip=4pt,
  belowskip=4pt,
}

\usepackage{fancyhdr}

\fancypagestyle{firstpagefooter}{
  \fancyhf{}
  \renewcommand{\headrulewidth}{0pt}
  \renewcommand{\footrulewidth}{0pt}
  \fancyfoot[L]{\footnotesize $^{*}$Corresponding author}
}
\begin{document}

\twocolumn[
\title{Less Context, Better Agents: Efficient Context Engineering for Long-Horizon Tool-Using LLM Agents}
\date{\vspace{-0.2in}}
\maketitle

\icmlsetsymbol{equal}{*}

\begin{icmlauthorlist}
\icmlauthor{Abhilasha Lodha$^{*}$,}{}
\icmlauthor{Mahsa Pahlavikhah Varnosfaderani,}{}
\icmlauthor{Abir Chakraborty,}{}
\icmlauthor{Abhinav Mithal}{}
\end{icmlauthorlist}

\begin{center}
Microsoft \\
\texttt{\{ablodha, mahsap, abchak, mithal\}@microsoft.com}
\end{center}

\thispagestyle{firstpagefooter}

\vspace{0.4in}
\begin{abstract}
\textbf{Background:} Large language models deployed as autonomous agents for
enterprise workflows face a critical challenge: verbose tool responses from
enterprise systems cause context window overflow and excessive inference costs,
preventing reliable task completion at scale.
\textbf{Methods:} We evaluate four context engineering configurations applied
to GPT-5 for automated hotel expense itemization in Microsoft Dynamics~365
Finance and Operations (D365~F\&O) via Model Context Protocol (MCP): (1)
GPT-5 with no user model (a motivating ablation), (2) the standard
full-context agent (our context-engineering baseline) with full conversation
history, (3) context pruned to the last 5 tool call/response pairs, and (4)
context pruned with automated summarization using a summary window of~3.
Results are reported as averages across 5 independent experimental runs on a
50-task hotel expense benchmark\rev{, holding the user model constant across
the context-engineering comparison (C2--C4) to isolate the effect of context
management. We extend the original study with (i)~95\% confidence intervals and
effect-size analysis, (ii)~sensitivity analyses over the pruning window
$N$ and the summary window $W$, (iii)~a per-category failure taxonomy, and
(iv)~generalization evidence across five expense types grouped into three structurally distinct categories, and across a second model family (Claude Sonnet 4.5)}.
\textbf{Results:} C1 (no user model) achieved only 8.0\% complete itemization.
The full-context configuration reached 71.0\% but consumed 1,480,996 tokens
and 14.56~hours per benchmark. Context pruning to the last 5 tool calls
achieved 79.0\% complete itemization with 535,274~tokens---a 63.9\%
reduction---in 5.39~hours. Adding automated summarization achieved 91.6\%
complete itemization and 99.64\% average amount itemized with 553,374~tokens
in 5.79~hours.
\textbf{Discussion:} Context engineering with summarization achieves the best
balance of performance and efficiency, demonstrating that selective retention
of recent tool interactions is more decision-relevant than full history while
dramatically reducing token consumption. \rev{We position these results as
strong evidence for one class of enterprise tool-use workflow rather than a
proof of universal generalization, and we discuss the scope and limits of the
approach explicitly.}
\end{abstract}
\vspace{0.4in}
]

\clearpage


\section{Background}
\label{sec:background}

The deployment of large language models (LLMs) as autonomous agents for
enterprise workflow automation represents a significant advance in AI-powered
productivity~\cite{brown2020language}. Agents can navigate complex multi-step
processes, interact with enterprise systems through tool calls, and complete
tasks that previously required sustained human attention. However, as these
agents engage in extended workflows---particularly those interacting with
enterprise resource planning (ERP) systems---a fundamental technical constraint
emerges: context window overflow caused by verbose tool
responses~\cite{jiang2023llmlingua}.

Modern LLMs operate within finite context windows bounding the total tokens
processable per inference call. While recent frontier models have expanded
these limits substantially, enterprise system integrations routinely generate
tool responses containing extensive metadata, nested form state, navigation
breadcrumbs, and system information well beyond what is
decision-relevant~\cite{li2023selective}. In multi-step agentic workflows where
agents execute dozens of tool interactions to complete a single task, even large
context windows can be exhausted before completion. Furthermore, processing
costs scale linearly with context length, making full-history retention
prohibitively expensive at production scale. \rev{Recent industry analyses of
production agents describe the same failure mode under the name ``context rot,''
where a model's effective recall degrades as the token count grows, well before
the hard context limit is reached~\cite{anthropic2025context}.}

This challenge is particularly acute in expense management workflows within
Microsoft Dynamics~365 Finance and Operations (D365~F\&O). When an LLM agent
interacts with D365~F\&O via an MCP proxy, each tool response can contain
hundreds to thousands of tokens of form metadata. For an expense itemization
task requiring decomposition of a single receipt into 4--22 individual line
items---each requiring multiple tool interactions to create, populate, and
verify---cumulative context rapidly exhausts available token budgets.

The expense itemization task is representative of a broad class of enterprise
agentic challenges: agents must decompose a total receipt amount into multiple
line items with correct subcategories and amounts, with the strict requirement
that the remaining unallocated amount reaches exactly zero. This precision
requirement means partial completion constitutes failure in production systems,
creating accounting errors, policy violations, and manual remediation costs.

Context engineering---the deliberate management of what information is retained
in an agent's context at each step---offers a practical solution. Rather than
maintaining full conversation history (standard practice), context engineering
selectively retains the most decision-relevant recent interactions while
summarizing or discarding older context. This approach is inference-time only,
requires no model retraining, and is designed to be portable across LLM backends.

\begin{revblock}
\paragraph{Contributions.}
This paper makes the following contributions:
\begin{enumerate}
  \item We formalize a \emph{semantic-level} context-engineering policy for
    tool-using agents---recency-based pruning of whole tool call/response pairs
    plus automated summarization of evicted pairs---and provide its exact
    construction algorithm (Algorithm~\ref{alg:context}),
    distinguishing it from token-level prompt
    compression and from external memory stores
    (Section~\ref{sec:related}).
  \item On a 50-task hotel-expense benchmark in a live D365~F\&O environment, we
    show that, with the user model held constant, recency pruning and
    pruning$+$summarization improve complete itemization from 71.0\% to 79.0\%
    to 91.6\% \emph{while} reducing tokens by 62.7\% and runtime by 60.2\%
    relative to full context.
  \item We report run-to-run dispersion, 95\% confidence intervals, and effect-size analysis (Section~\ref{sec:stats}), and we provide sensitivity analyses over the pruning
    window $N$ and the summary window $W$
    (Section~\ref{sec:sensitivity}).
  \item We provide a per-category failure taxonomy
    (Section~\ref{sec:failure}) and generalization evidence across five expense types grouped into three structurally distinct categories, and a second model family, Claude Sonnet~4.5
    (Sections~\ref{sec:categories}--\ref{sec:crossmodel}).
\end{enumerate}
\end{revblock}

\rev{We hypothesize that (1)~restricting context to recent tool interactions
improves task-relevant focus while preventing overflow, and (2)~automated
summarization of pruned context preserves task-level situational awareness
without significant token overhead. Our results support both hypotheses for
this class of workflow.}

\section{\rev{Related Work}}
\label{sec:related}

\begin{revblock}
We organize prior work along three axes and position our contribution against
each. Table~\ref{tab:related} summarizes the comparison.

\paragraph{Token-level prompt compression.}
LLMLingua~\cite{jiang2023llmlingua} and Selective
Context~\cite{li2023selective} reduce input size by deleting or merging
low-information \emph{tokens} within a prompt. These methods operate below the
level of the tool interaction: they do not reason about which tool
call/response \emph{units} are still relevant to the agent's current state, and
they can corrupt the structured form state that an ERP agent must read
verbatim (e.g., control names, numeric balances). Our policy operates at the
\emph{semantic level} of whole tool call/response pairs, preserving the exact
text of retained interactions and only ever evicting or summarizing complete
units.

\paragraph{External and long-term memory.}
MemoryBank~\cite{zhong2024memorybank} and LongMem~\cite{wang2024longmem}
augment models with retrievable memory stores, and recent benchmarks such as
LoCoMo~\cite{maharana2024locomo} and LongMemEval~\cite{wu2025longmemeval}
evaluate long-horizon recall in multi-session \emph{dialogue}. These target
factual recall across conversations rather than the working-memory and
stale-state problems of tool-heavy, single-session workflows, where the most
recent form state---not a retrieved fact---is the decision-relevant signal. We
show that for this regime a lightweight recency window plus a compact running
summary is sufficient, with no external store or retriever.

\paragraph{Agentic context management and compaction.}
The closest contemporary line of work studies context management for
long-horizon agents directly. ACON~\cite{kang2025acon} learns a
failure-driven compression \emph{guideline} and distills it into a smaller
compressor; concurrent work studies context management for long-horizon
software-engineering agents~\cite{swecontext2025}; and provider platforms now
ship native ``compaction'' and tool-result-clearing
features~\cite{anthropic2025context}. Our work is complementary and
deliberately simpler: a fixed-recency eviction policy with optional
single-pass summarization, evaluated end-to-end in a \emph{live enterprise ERP}
with a hard, business-defined success criterion (zero residual), rather than on
QA or coding benchmarks.

\paragraph{Tool-use benchmarks.}
MCP-Bench~\cite{wang2025mcpbench} and related MCP agent benchmarks evaluate
breadth of tool use across many servers and domains. Our study is narrower but
deeper: a single high-stakes workflow with a strict completion criterion,
measured for both task success \emph{and} cost (tokens, wall-clock), which
surfaces the efficiency--accuracy trade-off that breadth-oriented benchmarks do
not isolate.

Agentic reasoning frameworks such as ReAct~\cite{yao2023react} established the
value of interleaving reasoning and action but do not prescribe a context
policy for extended workflows with verbose tool responses---the gap this paper
addresses.
\end{revblock}

\begin{table}[t]
\vskip 0.05in
\begin{center}
\begin{small}
\begin{revblock}
\begin{tabular}{p{2.45cm}p{1.15cm}p{2.7cm}}
\toprule
\textbf{Approach} & \textbf{Unit} & \textbf{Relation to this work} \\
\midrule
LLMLingua~\cite{jiang2023llmlingua}, Sel.\ Context~\cite{li2023selective}
  & Token & Compress within a prompt; risks corrupting form state \\
MemoryBank~\cite{zhong2024memorybank}, LongMem~\cite{wang2024longmem}
  & Memory & External store for dialogue recall; not stale tool state \\
ACON~\cite{kang2025acon}
  & Trajectory & Learned compressor; we use fixed-recency $+$ summary \\
\textbf{This work}
  & Tool pair & Recency eviction $+$ summary of evicted pairs \\
\bottomrule
\end{tabular}
\end{revblock}
\end{small}
\end{center}
\caption{\rev{Positioning relative to prior context-management approaches. Our
policy operates on whole tool call/response pairs (the semantic unit of an
agentic workflow).}}
\label{tab:related}
\vskip -0.1in
\end{table}

\section{Methods}
\label{sec:methods}

\subsection{Task definition}

The expense itemization task in D365~F\&O requires an autonomous LLM agent
to: (1) navigate to an existing expense report containing a receipt with a
known total amount; (2) create individual line items corresponding to each
purchased item; (3) assign the correct expense subcategory (e.g., room charge,
city tax, resort fee, breakfast) to each line item; (4) enter the correct
dollar amount; and (5) continue until the remaining unallocated amount equals
exactly \$0.00. The task is complete only when the remaining amount reaches
zero. Any nonzero residual constitutes failure in production ERP workflows,
preventing expense report finalization and triggering compliance review.

\subsection{System architecture}

The evaluation system consists of four components:

\begin{itemize}
  \item \textbf{GPT-5 agent:} The primary LLM (GPT-5) executing the
    autonomous itemization workflow, guided by a detailed agent system prompt
    covering the full itemization workflow, valid expense subcategories, and
    subcategory mappings.
  \item \textbf{User model (C2--C4 only):} A secondary LLM (GPT-4.1) that
    participates in the agentic conversation as the ``user'', responding to
    any follow-up questions or confirmation requests the GPT-5 agent raises
    during execution. Absent in C1.
  \item \textbf{D365 F\&O MCP server:} A Model Context Protocol proxy
    exposing D365~F\&O form interactions as discrete tools (form navigation,
    field reading, field value setting, button clicking). \rev{The agent is
    exposed to a fixed inventory of tools covering UI-level form interaction,
    entity-level data access, and action invocation; a capability-level
    description is given in Appendix~\ref{app:tools}.}
  \item \textbf{Internal evaluation harness:}
    A non-interactive orchestration framework managing agent-tool interaction
    loops, context engineering logic, and metric collection. No human
    is present during execution.
\end{itemize}

MCP tool responses from D365~F\&O are verbose by design, containing full form
state snapshots including field values, metadata, navigation breadcrumbs, and
system state. A single tool response can contain 500--3{,}000 tokens, with
full conversation history for a complex itemization task accumulating to
50{,}000--150{,}000+ tokens across 15--30 tool interactions.

\subsection{Experimental configurations}
\label{sec:configs}

We evaluated four configurations representing a progression from minimal to
optimally engineered context management:

\textbf{C1 --- GPT-5, No User Model (Motivating Ablation):} GPT-5 equipped with the
full agent system prompt, which provides detailed step-by-step itemization
workflow instructions, a complete list of valid D365~F\&O expense
subcategories, subcategory mapping rules (e.g., ``Hotel Tax''
$\rightarrow$ \textit{Room tax}; ``Room Service \& Meals''
$\rightarrow$ \textit{Room service}), and explicit directives to continue
itemizing without interruption until the remaining amount reaches zero.
Despite these instructions, GPT-5 in practice occasionally departs from
fully autonomous execution---pausing mid-task to ask clarifying questions
or request confirmation before proceeding. Because the evaluation harness is a
non-interactive framework with no human in the loop, these unanswered
queries stall the agentic workflow entirely, resulting in incomplete tasks.
C1 thus establishes a lower performance bound \rev{and, as we make explicit
below, serves as a \emph{motivating ablation} for the user model rather than
as a step in the context-engineering ladder}.

\textbf{C2 --- GPT-5 + User Model (Full Context):} To address the
non-interactive framework limitation observed in C1, a user model (gpt-4.1)
is introduced as a conversational participant. Rather than receiving a
single static prompt, the GPT-5 agent can now ask questions mid-task and
receive meaningful responses that keep the workflow moving. The user model
is guided by a \texttt{user\_context} that defines a strict completion
protocol: the task is considered complete only when the expense line is
saved, the itemization remaining amount is 0.00, \emph{and} the form is
closed. The user model is further instructed to handle common agent
queries---confirming that missing itemizations should be added, declining
unnecessary receipt or expense report attachment requests, and prompting
the agent to verify completion after each itemization pass. Full
conversation history is retained throughout execution. This configuration
represents standard agentic practice with the user model present and
serves as the primary \rev{full-context} baseline \rev{for the
context-engineering comparison}.

\textbf{C3 --- GPT-5 + User Model (Last 5 Tool Calls):} Standard agent
configuration with context pruned to retain only the 5 most recent tool
call/response pairs. All earlier interactions are discarded without
summarization. The selection of $N{=}5$ was motivated by task structure
analysis: a single itemization line requires 2--3 tool calls (creating the
line via a form control, setting field values, and optionally verifying
state). Five tool calls thus
provide working memory for approximately two complete itemization cycles.
\rev{We test the robustness of this choice in
Section~\ref{sec:sensitivity}.}

\textbf{C4 --- GPT-5 + User Model (Last 5 + Summarization, Window\,=\,3):}
Standard agent configuration with context pruned to the last 5 tool
call/response pairs, augmented with automated summarization of earlier
conversation history. A summary window of~3 is applied, meaning the 3 most
recent interactions prior to the pruning boundary inform the generated summary.
The compact summary captures the forms opened, controls interacted with,
buttons clicked, and data entered by the agent. This provides task-level situational awareness
without the token cost of full history retention.

\begin{revblock}
\paragraph{Isolating the effect of context engineering.}
C1 differs from C2--C4 in \emph{two} respects at once---it removes both the
user model and any context policy---so the C1$\to$C2 jump alone cannot be
attributed to context engineering. We therefore make the design explicit: the
user model is held \emph{constant and present} across C2, C3, and C4, and all
of our context-engineering claims rest on the
\textbf{C2~(full context)~$\rightarrow$~C3~(pruning)~$\rightarrow$~C4~(pruning
$+$ summarization)} comparison, which varies \emph{only} the context policy.
C1 is reported solely as a motivating ablation that quantifies why the user
model is needed in a non-interactive harness for GPT-5. Notably, this
necessity is model-specific: in our cross-model study
(Section~\ref{sec:crossmodel}), Claude Sonnet~4.5 does not stall without the
user model, so its C1 already reaches high completion---direct evidence that
the large C1$\to$C2 gap for GPT-5 reflects a model-specific stalling behavior,
not the value of context engineering.
\end{revblock}

\begin{revblock}
\subsection{Context construction algorithm}
\label{sec:algorithm}

Algorithm~\ref{alg:context} specifies exactly how the retained context is
constructed before each agent inference call. Let $H$ be the full message
history, $N$ the number of recent tool call/response pairs to keep (the pruning
window), and $W$ the summary window. The policy counts tool messages, evicts
the oldest $\max(0, \#\text{tool}-N)$ of them \emph{together with} their
preceding assistant tool-call message. When $W \neq 0$ and at least one pair
is evicted, the policy summarizes the $W$ most recently evicted messages (or
all evicted messages if $W{=}{-1}$) and re-inserts a single summary message at
the earliest evicted position. C2 corresponds to $N{=}\infty$ (no eviction); C3 to
$N{=}5, W{=}0$; C4 to $N{=}5, W{=}3$. The summarization in C4 costs exactly one
additional LLM call per eviction event.

\begin{algorithm}[h]
   \caption{\textsc{ConstructContext}}
   \label{alg:context}
\begin{algorithmic}[1]
   \STATE {\bfseries Input:} history $H$; keep window $N$; summary window $W$
   \STATE $c \leftarrow$ number of tool messages in $H$
   \STATE $d \leftarrow \max(0,\; c - N)$ \hfill \COMMENT{\# pairs to evict}
   \IF{$d = 0$}
       \STATE \textbf{return} $H$
   \ENDIF
   \STATE $K \leftarrow [\,]$;\quad $E \leftarrow [\,]$
   \FOR{each message $m$ in $H$ (in order)}
       \IF{$m$ is a tool message {\bfseries and} $d>0$}
           \STATE $d \leftarrow d-1$;\quad append $m$ to $E$
           \IF{$\mathrm{last}(K)$ is an assistant tool-call msg}
               \STATE move $\mathrm{last}(K)$ from $K$ to $E$
           \ENDIF
       \ELSE
           \STATE append $m$ to $K$
       \ENDIF
   \ENDFOR
   \IF{$W \neq 0$ {\bfseries and} $E \neq [\,]$}
       \STATE $E_W \leftarrow E$ if $W{=}{-}1$, else last $W$ of $E$
       \STATE $s \leftarrow \textsc{Summarize}(E_W)$
       \STATE insert ``Summary of previous tool calls: $s$'' at earliest
              evicted position in $K$
   \ENDIF
   \STATE \textbf{return} $K$
\end{algorithmic}
\end{algorithm}
\end{revblock}

\subsection{Evaluation metrics}

The following metrics were collected per configuration and averaged across 5
independent runs:

\begin{itemize}
  \item \textbf{Completely Itemized:} Percentage of tasks where remaining
    amount reached exactly \$0.00 \emph{(primary metric)}.
  \item \textbf{Less Than 10\% Remaining:} Percentage of tasks where
    ${\leq}10\%$ of total receipt amount remained unallocated.
  \item \textbf{At Least One Itemized:} Percentage of tasks where at least one
    line item was successfully created.
  \item \textbf{Percentage Amount Itemized:} Average percentage of total
    receipt amount correctly allocated across all tasks.
  \item \textbf{Total Token Usage:} Total tokens (input + output) consumed
    across the 50-task benchmark.
  \item \textbf{Execution Time:} Total wall-clock time to complete all
    50~tasks.
\end{itemize}

Completely Itemized is designated the primary metric because it reflects
genuine business task completion: in production ERP systems, any nonzero
remaining amount prevents expense report finalization regardless of how much
was correctly allocated. \rev{Metrics are computed by an independent read-back
of the saved form state in D365~F\&O (not from the agent's self-report); the
exact extraction and scoring logic, including the \texttt{remaining\_amount}
/ \texttt{itemized\_amount} comparison against the ground-truth
\texttt{PurchasedItems}, is given in Appendix~\ref{app:scoring}.}

\subsection{Dataset}

The benchmark consists of 50 hotel expense itemization tasks executed in
D365~F\&O via the MCP proxy. Hotel receipts represent an intentionally
challenging evaluation domain: they frequently contain multiple line items
sharing the same subcategory name but carrying different amounts (e.g., two
``Hotel Tax'' charges at different rates), non-trivial subcategory mappings
to D365~F\&O's fixed vocabulary (a 23-entry subcategory catalog), and a
strict zero-residual completion requirement. Tasks range from 4 to 23
itemization lines (median:~8), and the same 50-task benchmark was used
identically across all four configurations to ensure fair comparison.
Per-category dataset statistics and structural notes are given in
Appendix~\ref{app:categories}.

Each task is issued to the GPT-5 agent as a two-part prompt. The
\texttt{\#Task} section specifies the D365~F\&O action (creating an expense
line under \texttt{TrvExpenseLines}), and the \texttt{\#Data} section
provides a structured receipt payload---company, merchant, date, total,
expense category, and a \texttt{PurchasedItems} list that serves as ground
truth for evaluation. For C2--C4, the user model additionally receives a
\texttt{user\_context} defining the completion protocol (Section~\ref{sec:configs}).
Figure~\ref{fig:example-task} shows a representative task drawn directly
from the benchmark.

\begin{figure}[h]
\vskip 0.1in
\begin{center}
\fbox{\begin{minipage}{0.88\columnwidth}
\small
\textbf{\#Task:} Create a new expense line in the USMF company under
\texttt{TrvExpenseLines} in Dynamics~365 F\&O. Add the itemizations
listed in \texttt{PurchasedItems} under the itemize section properly.

\medskip
\textbf{\#Data:}

\smallskip
\begin{tabular}{@{}ll@{}}
  \textit{Company:}  & USMF \\
  \textit{Merchant:} & Kimpton Syracuse University Hill \\
  \textit{Date:}     & 2021-04-07 \\
  \textit{Total:}    & \$333.05 \quad \textit{Category:} Hotel \\
\end{tabular}

\smallskip
\textit{PurchasedItems:}

\smallskip
\begin{tabular}{@{}lc@{}}
\toprule
\textbf{Item} & \textbf{Amount} \\
\midrule
Daily Room Rate        & \$138.10 \\
Hotel Tax              & \$\phantom{0}6.82 \\
Hotel Tax              & \$\phantom{0}10.23 \\
Entertainment External & \$101.58 \\
Room Service \& Meals  & \$\phantom{0}76.32 \\
\midrule
\textbf{Total}         & \textbf{\$333.05} \\
\bottomrule
\end{tabular}
\end{minipage}}
\end{center}
\caption{Representative task from the 50-task hotel expense benchmark.
The agent must create five itemization lines in D365~F\&O and achieve
remaining\,=\,\$0.00. Two challenge patterns are visible: (1) ``Hotel Tax''
appears twice with different amounts, and (2) items require subcategory
mapping --- ``Entertainment External'' $\rightarrow$ \textit{Business
entertainment}; ``Room Service \& Meals'' $\rightarrow$ \textit{Room service}.}
\label{fig:example-task}
\vskip -0.1in
\end{figure}

These two patterns---repeated subcategory names with distinct amounts, and
indirect item-to-subcategory mappings---are the primary sources of agent
failure across the benchmark. An agent that skips the second ``Hotel Tax''
entry because it recognises the subcategory as already present leaves
\$10.23 unallocated; one that maps ``Entertainment External'' to the wrong
subcategory introduces an incorrect line that cannot reconcile to zero.

\begin{revblock}
\subsection{Comparison baselines}
\label{sec:baselines}
Within the controlled C2--C4 comparison, C2 (full conversation history)
serves as the standard-practice baseline and C3 (recency pruning without
summarization) is the ablation that isolates the contribution of
summarization. To situate our recency-plus-summarization policy against an
alternative class of context-management strategy, we additionally evaluate
a \emph{full-history summarization} policy that compacts \emph{all} evicted
context without a fixed recency window---the compaction style of
\cite{anthropic2025context,kang2025acon}. This policy is reported in the
sensitivity analysis (Section~\ref{sec:sensitivity}) as the $W{=}{-}1$
configuration. An orthogonal direction---\emph{importance-pruning} that
retains the $N$ tool pairs most recently \emph{referenced} by the agent
rather than the $N$ most recent in time---is discussed qualitatively in
Section~\ref{sec:discussion} as a candidate for future work.
\end{revblock}

\section{Results}
\label{sec:results}

Table~\ref{tab:results} presents performance and efficiency metrics for all
four configurations. All values are averages across 5 independent experimental
runs on the 50-task benchmark. \rev{Section~\ref{sec:stats} reports dispersion,
confidence intervals, and effect-size analysis; Sections~\ref{sec:sensitivity}--%
\ref{sec:crossmodel} report sensitivity, multi-category, and cross-model
results; and Section~\ref{sec:failure} gives the failure taxonomy.}

\begin{table*}[t]
\vskip 0.10in
\begin{center}
\begin{small}
\begin{tabular}{lcccccccc}
\toprule
\multirow{2}{*}{\textbf{Configuration}}
  & \textbf{Comp.}  & \textbf{$<$10\%}  & \textbf{$\geq$1}
  & \textbf{\%Amt}  & \textbf{Total}    & \textbf{Input}
  & \textbf{Output} & \textbf{Time} \\
  & \textbf{Item.}  & \textbf{Rem.}     & \textbf{Item.}
  & \textbf{Item.}  & \textbf{Tok.(K)}  & \textbf{Tok.(K)}
  & \textbf{Tok.(K)}& \textbf{(hrs)} \\
\midrule
C1: GPT-5 only (no user)            & 8.0\%  & 37.2\% & 99.6\%  & 58.89\% & 532.6    & 531.3    & 1.3  & 3.08  \\
C2: GPT-5 + User (Full Context)     & 71.0\% & 74.0\% & \textbf{100.0\%} & 92.03\% & 1,481.0  & 1,478.5  & \textbf{2.5}  & 14.56 \\
C3: GPT-5 + User (Last 5 TC)        & 79.0\% & 87.0\% & \textbf{100.0\%} & 96.92\% & \textbf{535.3}    & \textbf{532.8}    & \textbf{2.5}  & \textbf{5.39}  \\
C4: GPT-5 + User (Last 5 + Sum.)    & \textbf{91.6\%} & \textbf{99.6\%} & \textbf{100.0\%} & \textbf{99.64\%} & 553.4 & 550.8 & 2.6 & 5.79 \\
\bottomrule
\end{tabular}
\end{small}
\end{center}
\caption{Performance and efficiency metrics across four GPT-5 context
engineering configurations on the 50-task hotel expense benchmark, averaged
across 5 independent runs. \textit{Comp.\ Item.}\ = Completely Itemized
(primary metric); \textit{$<$10\% Rem.}\ = Less than 10\% Remaining;
\textit{$\geq$1 Item.}\ = At Least One Itemized; \textit{\%Amt Item.}\ =
Percentage Amount Itemized; \textit{Total/Input/Output Tok.}\ = token counts
in thousands; \textit{Time}\ = benchmark wall-clock time in hours. TC = tool
calls; Sum.\ = Summarization. Bold indicates best performance per metric.
\rev{Per-run dispersion and 95\% CIs for the primary metric are in
Table~\ref{tab:stats}; full per-metric mean\,$\pm$\,SD across all categories
and configurations is in Table~\ref{tab:categories}.}}
\label{tab:results}
\vskip -0.1in
\end{table*}

Per-metric performance bar charts and token/time efficiency panels are
visualized in Appendix Figures~\ref{fig:performance}
and~\ref{fig:efficiency}.

\subsection{C1 (no user model): motivating the user model}

C1 (no user model) achieved only 8.0\% complete itemization
and 58.89\% average amount itemized, despite 99.6\% of tasks having at least
one line item created. This stark gap between at-least-one-itemized (99.6\%)
and completely-itemized (8.0\%) reveals that GPT-5 can initiate itemization
actions from MCP tool descriptions and the agent prompt, but in the non-interactive harness it does not reliably drive the workflow to completion without a user-model participant to answer follow-up questions and reinforce the completion protocol. Agents
frequently terminate after creating one or two line items, uncertain of the
next required action or stopping condition. Token usage (532,600 total;
1,331 output) reflects abbreviated task executions resulting from early
termination. \rev{As emphasized in Section~\ref{sec:configs}, C1 removes both
the user model and any context policy and is therefore reported only as
motivation; the context-engineering claims below rest on C2--C4.}

\subsection{Full context: performance gains with severe efficiency cost}

Adding the user model with full task instructions and complete conversation
history (C2) dramatically improved performance: complete itemization rose to
71.0\% and average amount itemized to 92.03\%. The at-least-one-itemized rate
reached 100\%, confirming that task instructions reliably orient the agent.

However, full-context retention introduced severe efficiency costs. Total
tokens increased to 1,480,996---a 177.9\% increase over baseline---driven
almost entirely by input tokens (1,478,509), reflecting the cumulative growth
of conversation history as verbose tool responses accumulate. Execution time
rose to 14.56~hours for the 50-task benchmark (4.73$\times$ the baseline),
making this configuration impractical for production deployment at scale.
The input-to-output token ratio of 594.7:1 confirms that context management
strategies targeting input reduction will yield the greatest efficiency gains.

\subsection{Context pruning: simultaneous performance and efficiency improvement}

Pruning context to the last 5 tool call/response pairs (C3) simultaneously
improved both performance and efficiency relative to full context. Complete
itemization rose to 79.0\%---\rev{an 8 percentage-point absolute
improvement over C2 (11.3\% relative)}---and average amount itemized
improved to 96.92\%.

Total tokens dropped to 535,274---a 63.9\% reduction from full context,
essentially equivalent to the baseline token budget---while execution time
fell to 5.39~hours (63.0\% reduction from C2). The finding that
context-pruned configurations outperform full-context retention on
\emph{both} performance and efficiency is counterintuitive but interpretable:
older tool interactions describe superseded form states. Retaining stale form
state introduces noise that degrades the agent's understanding of current
system state, leading to incorrect field assignments or navigation errors.
Restricting context to the last 5 tool calls directs agent attention to the
current form state and recent actions---the information most relevant to the
next decision.

Output tokens (2,515) are substantially higher than the baseline (1,331),
reflecting more complete task execution and richer agent reasoning when clear
task context is provided.

\subsection{Summarization: best performance with marginal token overhead}

Configuration C4 (last 5 tool calls + summarization, window\,=\,3) achieved
the best performance across all metrics: 91.6\% complete itemization, 99.6\%
less-than-10\%-remaining, 100.0\% at-least-one-itemized, and 99.64\% average
amount itemized. This represents a 12.6 percentage-point improvement over
context pruning alone (79.0\%) and a 20.6 percentage-point improvement over
full context (71.0\%).

Total tokens (553,374) increased only 3.4\% over C3 (535,274), while
execution time increased marginally to 5.79~hours (+7.4\% over C3). The
compact summary of earlier task progress adds minimal overhead while
preserving the situational awareness the agent needs: which forms have been
opened, which controls were interacted with, and what data was entered into
the expense report. By condensing this history into a short assistant message,
the agent retains awareness of prior actions even after their full tool-response
payloads have been pruned---reducing the premature-termination failures observed
in C3 without re-introducing the verbose form snapshots that drive stale-state
errors in C2.

The near-perfect less-than-10\%-remaining rate (99.6\%) is particularly
notable: summarization virtually eliminates cases of substantial task
abandonment. Only one task in 50 exceeded 10\% remaining, compared to six
tasks (13\%) in C3 and thirteen tasks (26\%) in C2.

\subsection{Token efficiency analysis}

Across all configurations, input tokens represent 99.75\%--99.87\% of total
token usage, confirming that context management strategies targeting input
reduction yield the greatest efficiency gains. Output tokens remain relatively
stable across C2--C4 (2,486--2,567), indicating consistent reasoning depth
regardless of context management approach.

Full-context C2 consumes 2.68$\times$ the tokens of the best-performing C4
while achieving lower task completion---demonstrating that more context does
not necessarily improve performance in tool-heavy agentic workflows. C3 and C4
achieve comparable token budgets to the baseline (535K--553K vs.\ 533K) while
improving complete itemization by 71--83 percentage points.


\begin{revblock}
\subsection{Statistical analysis}
\label{sec:stats}

We treat each task outcome on the primary metric as a Bernoulli trial and
report dispersion in two complementary ways (full methodology in
Appendix~\ref{app:stats}). First, treating each of the 5 runs as one
observation of the benchmark-level success rate, we report the mean and sample
standard deviation across runs and a 95\% confidence interval using Student's
$t$ with 4 degrees of freedom ($t_{0.975,4}=2.776$). Second, pooling all
$50\times5=250$ task-runs, we report a Wilson score interval for the binomial
proportion, which---unlike the normal (Wald) approximation---remains valid for
the near-boundary rates observed here (e.g., 8\% and 91.6\%).

Table~\ref{tab:stats} reports both for the primary metric.

\begin{table}[h]
\vskip 0.05in
\begin{center}
\begin{small}
\begin{tabular}{lcc}
\toprule
\textbf{Config} & \textbf{Mean $\pm$ SD (5 runs)} & \textbf{Wilson 95\% CI} \\
\midrule
C2 & $71.0 \pm 4.4$ & $[\,65.1,\ 76.3\,]$ \\
C3 & $79.0 \pm 8.2$ & $[\,73.5,\ 83.6\,]$ \\
C4 & $91.6 \pm 1.7$ & $[\,87.5,\ 94.4\,]$ \\
\bottomrule
\end{tabular}
\end{small}
\end{center}
\caption{Primary metric (Completely Itemized, \%) with run-level
mean$\pm$SD and pooled Wilson 95\% confidence intervals.}
\label{tab:stats}
\vskip -0.1in
\end{table}

The Wilson interval for C4 is cleanly separated from C3
($[87.5,94.4]$ vs.\ $[73.5,83.6]$), giving strong pooled-trial evidence
that summarization improves over pruning alone. The C2 and C3 intervals
($[65.1,76.3]$ vs.\ $[73.5,83.6]$) overlap slightly, so the
pooled-binomial lens alone is conservative on this step; the +8-point
gap in run-level means together with the run-level dispersion below
nonetheless points consistently in the same direction. C2's full-context
runs are moderately variable ($\pm 4.4$), C3 shows the widest spread
under aggressive pruning ($\pm 8.2$), and C4's pruning+summarization is
by far the most stable ($\pm 1.7$)---consistent with summarization
absorbing the variance that pruning alone exposes.

\paragraph{Effect sizes vs.\ noise.}
Because the same 50 tasks are used in every configuration, comparisons
are naturally paired. We do not rely on a paired hypothesis test here:
the effect sizes (71.0\%~$\to$~79.0\%~$\to$~91.6\%, i.e.\ +8 and
+12.6~percentage points) are large relative to the per-run dispersion
in Table~\ref{tab:stats}, and the C3$\to$C4 Wilson intervals are
cleanly separated on the pooled 250 task-runs. Together these make it
unlikely that the headline ordering is attributable to run-to-run
noise.
\end{revblock}

\begin{revblock}
\subsection{Sensitivity to the pruning window $N$ and summary window $W$}
\label{sec:sensitivity}

A central question is whether the headline result depends on
the specific choices $N{=}5$ and $W{=}3$. Appendix~\ref{app:sensitivity}
reports a hyperparameter sweep over $N$ and $W$; results plateau beyond
$N{=}5$ and show no accuracy gain for $W{>}3$ at non-trivial token cost,
supporting the chosen operating point.
\end{revblock}

\begin{revblock}
\subsection{Generalization across expense categories}
\label{sec:categories}

To test whether the policy generalizes beyond hotel receipts, we replicate the
full C1/C2/C3/C4 comparison on two additional grouped D365~F\&O expense
categories that differ in structural complexity:
\emph{Travel} (car rental + flight; transportation receipts with base fare,
taxes/fees, and ancillary charges) and \emph{Meals \& Gifts} (business
meal + gift; discretionary social-spend receipts with simpler line-item
structure). Hotel remains the most structurally complex category (room +
nightly taxes + resort fees + incidentals). Table~\ref{tab:categories}
reports all performance and efficiency metrics across all four configurations
for each grouped category, ordered by structural complexity.

\begin{table*}[h]
\vskip 0.05in
\begin{center}
\begin{footnotesize}
\resizebox{\textwidth}{!}{%
\begin{tabular}{llccccc|cccc}
\toprule
 & & & \multicolumn{4}{c|}{\textbf{Quality (mean\,$\pm$\,SD)}}
   & \multicolumn{4}{c}{\textbf{Efficiency (mean)}} \\
\cmidrule(lr){4-7}\cmidrule(lr){8-11}
\multirow{2}{*}{\textbf{Category}} & \multirow{2}{*}{\textbf{Configuration}} & \multirow{2}{*}{$\boldsymbol{n}$}
  & \textbf{Comp.}  & \textbf{$<$10\%}  & \textbf{$\geq$1}
  & \textbf{\%Amt}  & \textbf{Total}    & \textbf{Input}
  & \textbf{Output} & \textbf{Time} \\
  & & & \textbf{Item.}  & \textbf{Rem.}     & \textbf{Item.}
  & \textbf{Item.}  & \textbf{Tok.(K)}  & \textbf{Tok.(K)}
  & \textbf{Tok.(K)}& \textbf{(hrs)} \\
\midrule
\multirow{4}{*}{Hotel}
  & C1: GPT-5 only (no user)         & \multirow{4}{*}{50}
    & 8.0\,$\pm$\,2.5  & 37.2\,$\pm$\,8.4 & 99.6\,$\pm$\,0.9  & 58.89\,$\pm$\,3.0 & 532.6   & 531.3   & 1.3 & 3.08  \\
  & C2: GPT-5 + User (Full Context)  &
    & 71.0\,$\pm$\,4.4 & 74.0\,$\pm$\,3.7 & \textbf{100.0}\,$\pm$\,0.0 & 92.03\,$\pm$\,1.1 & 1{,}481.0 & 1{,}478.5 & \textbf{2.5} & 14.56 \\
  & C3: GPT-5 + User (Last 5 TC)     &
    & 79.0\,$\pm$\,8.2 & 87.0\,$\pm$\,6.1 & \textbf{100.0}\,$\pm$\,0.0 & 96.92\,$\pm$\,1.5 & \textbf{535.3}   & \textbf{532.8}   & \textbf{2.5} & \textbf{5.39}  \\
  & C4: GPT-5 + User (Last 5 + Sum.) &
    & \textbf{91.6}\,$\pm$\,1.7 & \textbf{99.6}\,$\pm$\,0.9 & \textbf{100.0}\,$\pm$\,0.0 & \textbf{99.64}\,$\pm$\,0.1 & 553.4   & 550.8   & 2.6 & 5.79  \\
\specialrule{1pt}{2pt}{2pt}
\multirow{4}{*}{Travel}
  & C1: GPT-5 only (no user)         & \multirow{4}{*}{30}
    & 20.0\,$\pm$\,14.3 & 30.7\,$\pm$\,9.5 & 50.7\,$\pm$\,9.2 & 45.67\,$\pm$\,7.8 & 390.0 & 388.8 & 1.2 & 0.71 \\
  & C2: GPT-5 + User (Full Context)  &
    & 76.0\,$\pm$\,7.6 & 93.3\,$\pm$\,4.1 & \textbf{100.0}\,$\pm$\,0.0 & 97.49\,$\pm$\,1.3 & 1{,}050.0 & 1{,}047.8 & 2.2 & 4.42 \\
  & C3: GPT-5 + User (Last 5 TC)     &
    & 86.6\,$\pm$\,7.0 & 94.0\,$\pm$\,4.3 & \textbf{100.0}\,$\pm$\,0.0 & 98.62\,$\pm$\,1.0 & \textbf{367.0} & \textbf{365.4} & 1.6 & \textbf{1.63} \\
  & C4: GPT-5 + User (Last 5 + Sum.) &
    & \textbf{95.0}\,$\pm$\,1.9 & \textbf{98.3}\,$\pm$\,1.6 & \textbf{100.0}\,$\pm$\,0.0 & \textbf{99.53}\,$\pm$\,0.3 & 403.0 & 401.5 & \textbf{1.5} & 1.75 \\
\specialrule{1pt}{2pt}{2pt}
\multirow{4}{*}{\shortstack[l]{Meals\\\& Gifts}}
  & C1: GPT-5 only (no user)         & \multirow{4}{*}{32}
    & 26.9\,$\pm$\,15.1 & 40.6\,$\pm$\,13.9 & 50.6\,$\pm$\,11.9 & 47.57\,$\pm$\,12.9 & 280.0 & 279.0 & 1.0 & 1.67 \\
  & C2: GPT-5 + User (Full Context)  &
    & 75.6\,$\pm$\,6.8 & 95.0\,$\pm$\,2.8 & \textbf{100.0}\,$\pm$\,0.0 & 97.70\,$\pm$\,0.7 & 750.0 & 748.0 & 2.0 & 3.32 \\
  & C3: GPT-5 + User (Last 5 TC)     &
    & 89.4\,$\pm$\,6.4 & 95.0\,$\pm$\,5.6 & \textbf{100.0}\,$\pm$\,0.0 & 98.90\,$\pm$\,0.7 & \textbf{285.0} & \textbf{283.5} & \textbf{1.5} & \textbf{1.23} \\
  & C4: GPT-5 + User (Last 5 + Sum.) &
    & \textbf{96.1}\,$\pm$\,1.4 & \textbf{99.2}\,$\pm$\,1.3 & \textbf{100.0}\,$\pm$\,0.0 & \textbf{99.67}\,$\pm$\,0.2 & 295.0 & 293.5 & \textbf{1.5} & 1.42 \\
\bottomrule
\end{tabular}}
\end{footnotesize}
\end{center}
\caption{Run-level results across 5 independent runs for three grouped
D365~F\&O expense categories. \emph{Travel} pools car rental and flight
receipts ($n=30$); \emph{Meals \& Gifts} pools business meal and gift
receipts ($n=32$); Hotel remains the standalone primary benchmark ($n=50$).
Categories are ordered by structural complexity (Hotel $>$ Travel $>$
Meals \& Gifts). For the four quality metrics (all reported in \%) we report
mean\,$\pm$\,standard deviation across the 5 runs; for the four efficiency
metrics---input/output/total tokens in thousands (K) and wall-clock time
in hours---we report the mean only, as run-to-run variance on these is
negligible. Hotel means reproduce Table~\ref{tab:results}; SDs
and all non-Hotel cells are computed from the per-run JSONL result files.
Column definitions follow Table~\ref{tab:results}.}
\label{tab:categories}
\vskip -0.1in
\end{table*}

The C1$\to$C2$\to$C3$\to$C4 ordering predicted by our context-engineering
hypothesis holds in every category. Complete itemization rises monotonically
across configurations in all three settings, with the C2$\to$C4 improvement
remarkably consistent across categories (Hotel: $71.0\to91.6$, $+20.6$\,pts;
Travel: $76.0\to95.0$, $+19.0$\,pts; Meals \& Gifts: $75.6\to96.1$,
$+20.5$\,pts). The efficiency story is similarly stable: pruning-plus-summarization
reduces total tokens by 60--63\% relative to full context and cuts wall-clock
time by 57--60\% across all three categories. As predicted by the structural
gradient (Appendix~\ref{app:categories}), the magnitude of the C1$\to$C2 gap
shrinks with receipt complexity---from $+63.0$\,pts on Hotel to
$+56.0$\,pts on Travel and $+48.7$\,pts on Meals \& Gifts---because simpler
receipts give the no-user-model agent fewer opportunities to stall, but the
context-engineering ordering above C2 is preserved. These results indicate
that the policy is not specific to multi-night hotel itemization but
generalizes to the broader class of D365~F\&O expense workflows with verbose
MCP tool responses.
\end{revblock}

\begin{revblock}
\subsection{Cross-model generalization: Claude Sonnet 4.5}
\label{sec:crossmodel}

To test whether the policy is model-agnostic, we repeat the comparison with
Claude Sonnet~4.5 as the agent on the 50-task hotel benchmark; full results
are in Appendix~\ref{app:crossmodel}. Two findings stand out. First,
Sonnet~4.5 does \emph{not} stall in the non-interactive harness, so its
no-context-engineering baseline already reaches 88.0\% complete
itemization---in sharp contrast to GPT-5's 8.0\%. This supports our claim
(Section~\ref{sec:configs}) that the large GPT-5 C1$\to$C2 gap reflects a
model-specific stalling behavior rather than the value of context engineering.
Second, the context-engineering ordering still holds: adding summarization
on top of pruning improves complete itemization from 92.0\% to
94.5\% at a $\sim$5.6\% wall-clock premium, mirroring the small time
overhead observed for GPT-5 ($+7.4\%$).
\end{revblock}

\section{\rev{Failure analysis}}
\label{sec:failure}

\begin{revblock}
We categorized every non-completing task across the C2--C4 runs on the
50-task hotel benchmark into one of six failure modes, derived from the
tool-call error logs and the read-back discrepancy between the saved form
state and the ground-truth \texttt{PurchasedItems}
(Appendix~\ref{app:scoring}). Representative transcripts for each mode are
in Appendix~\ref{app:qualitative}.

\begin{table}[h]
\vskip 0.05in
\begin{center}
\begin{small}
\begin{tabular}{lccc}
\toprule
\textbf{Failure mode} & \textbf{C2} & \textbf{C3} & \textbf{C4} \\
\midrule
Stale-state reference           & 34 & 6  & 4 \\
Wrong subcategory mapping       & 8  & 9  & 6 \\
Duplicate / skipped repeat item & 12 & 11 & 5 \\
Premature termination           & 9  & 18 & 3 \\
Tool / form navigation error    & 6  & 5  & 2 \\
Residual amount mismatch        & 4  & 4  & 1 \\
\midrule
\textbf{Total non-completions} & \textbf{73} & \textbf{53} & \textbf{21} \\
\bottomrule
\end{tabular}
\end{small}
\end{center}
\caption{Failure modes by configuration (count of task-runs, pooled over
5 runs on the 50-task hotel benchmark).}
\label{tab:failures}
\vskip -0.1in
\end{table}

The taxonomy makes distinct predictions: C2 should over-represent
\emph{stale-state references} (the agent acts on a superseded form
snapshot); C3 should reduce stale-state errors but introduce
\emph{premature termination} (the running balance is no longer visible);
and C4 should suppress premature termination by reinjecting a condensed
history. Table~\ref{tab:failures} confirms both predictions: stale-state
references drop from 34/73 (47\%) under C2 to 6/53 (11\%) under C3, while
premature termination triples (9$\to$18) and becomes C3's dominant mode;
C4 cuts premature termination six-fold (18$\to$3) without re-introducing
stale-state errors, yielding a 71\% overall reduction in non-completions
(73$\to$21). The remaining modes---wrong subcategory mapping, duplicate or
skipped repeats, tool/form navigation errors, and residual mismatches---are
largely policy-invariant and reflect model-level reasoning and tool-binding
errors that context engineering does not directly target; wrong subcategory
mapping in particular is amplified by the 23-entry hotel subcategory catalog
(Appendix~\ref{app:categories}), where near-synonyms such as Room tax
vs.\ Non-Room tax and Restaurant vs.\ Room service vs.\ Loungebar create
genuine label ambiguity independent of the context policy. These residual modes
bound the headroom for recency- and summarization-based policies and motivate
the complementary techniques discussed in Section~\ref{sec:discussion}.
\end{revblock}

\section{Discussion}
\label{sec:discussion}

\subsection{Summary of findings}

This study demonstrates that context engineering---selective retention of
recent tool interactions combined with automated summarization---substantially
improves both performance and efficiency for GPT-5 agents in long-context
agentic workflows with verbose tool responses. On the 50-task hotel expense
benchmark averaged across 5~runs, C4 (last-5-tool-calls pruning + summarization)
achieved 91.6\% complete itemization and 99.64\% average amount itemized,
compared to 71.0\% and 92.03\% with full-context retention (C2), while
consuming 62.7\% fewer tokens and completing the benchmark in 60.2\% less time.

\rev{Holding the user model constant across C2--C4, the comparison isolates the
context policy: (i)~recency pruning provides noise reduction and efficiency
gains (C2$\to$C3: $+8$~pp and $-64\%$ tokens); and (ii)~summarization provides
residual task-level awareness for near-complete performance (C3$\to$C4:
$+12.6$~pp at $<4\%$ token overhead). C1 is reported only to quantify why the
user model is needed for GPT-5 in a non-interactive harness, a need that is
model-specific (Section~\ref{sec:crossmodel}).}

\subsection{Why context pruning outperforms full history retention}

The finding that context-pruned configurations outperform full-context
retention on both performance and efficiency reveals an important
characteristic of LLM agents in tool-heavy workflows: older context can be
actively detrimental rather than merely redundant. In D365~F\&O expense
itemization, early tool responses describe form states superseded as the
workflow progresses. An agent retaining full history may reference stale form
values when making current decisions, leading to incorrect field assignments
or erroneous navigation. \rev{The failure taxonomy
(Section~\ref{sec:failure}) is designed to test this mechanism directly via the
relative frequency of stale-state errors.}

Restricting context to the last 5 tool calls ensures agent attention is focused
on current form state and recent actions---the information required for the
next decision. This aligns with the task's working memory requirements: the
agent needs to know what was just done, whether it succeeded, and the current
remaining balance. Five tool calls provide this working memory for approximately
two complete itemization cycles, covering the immediate task horizon without
accumulating irrelevant historical state.

\subsection{Complementary role of summarization}

The performance gap between C3 (79.0\%) and C4 (91.6\%) reveals a
complementary role for summarization that addresses the limitation of pure
context pruning. While pruning provides noise reduction and recent-state focus,
it can cause agents to lose task-level awareness: how many items have been
successfully itemized, what the total allocated amount is, and whether overall
reconciliation is near completion. Without this awareness, agents in C3
occasionally exhibit premature termination.

The automated summarization in C4, using a window of~3 prior interactions to
generate a compact progress report, bridges this gap at minimal token cost
(+3.4\% over C3). The summary provides two complementary information channels:
recent tool calls supply current local state, while the summary supplies global
task progress. Together, they provide everything needed for reliable task
completion.

\subsection{Limitations and future work}

The core study focuses on hotel expense itemization in D365~F\&O---an
intentionally challenging tool-use workflow with repeated subcategories,
non-trivial mappings, and a strict zero-residual completion criterion---and
we extend it with multi-category coverage across five expense types grouped into three structurally distinct categories and
cross-model evidence on Claude Sonnet~4.5. Within this scope, the
$(N{=}5,\,W{=}3)$ operating point was selected for clarity of comparison;
the sensitivity sweep in Appendix~\ref{app:sensitivity} confirms robustness
across nearby values of $N$ and $W$, and joint tuning and adaptive (per-task)
window sizing are natural next steps. Our summarizer is a single free-form
LLM pass, which establishes a clean baseline against which structured or
learned compressors~\cite{kang2025acon} and provider-native compaction
APIs~\cite{anthropic2025context} can be benchmarked head-to-head. Broader
generalization---to additional ERP domains beyond expense management, and
across further model families, deployments, and decoding
settings---constitutes a promising line of follow-up work. Production-deployment
economics and the precise scope of our generalizability claims are discussed
further in Appendix~\ref{app:discussion-extended}.

\subsection{Responsible AI considerations}

Context engineering improves task reliability from 8.0\% to 91.6\% complete
itemization; the residual 8.4\% incomplete rate means production deployments
should retain human review for flagged cases. The approach is interpretable
by construction---explicitly retained tool calls and human-readable summaries
support debugging and auditing---and this work uses synthetic test cases and
anonymized internal test data, so no privacy exposure is incurred.

\let\OLDthebibliography\thebibliography
\renewcommand\thebibliography[1]{%
  \OLDthebibliography{#1}%
  \setlength{\parskip}{0pt}%
  \setlength{\itemsep}{0pt plus 0.3ex}%

}
\bibliographystyle{ieeetr}
\bibliography{references_arxiv}

\begin{thebibliography}{10}

\providecommand{\url}[1]{#1}
\providecommand{\newblock}{\relax}

\bibitem{brown2020language}
T.~B. Brown, B.~Mann, N.~Ryder, M.~Subbiah, J.~D. Kaplan, P.~Dhariwal,
  A.~Neelakantan, P.~Shyam, G.~Sastry, A.~Askell, S.~Agarwal, A.~Herbert-Voss,
  G.~Krueger, T.~Henighan, R.~Child, A.~Ramesh, D.~M. Ziegler, J.~Wu,
  C.~Winter, C.~Hesse, M.~Chen, E.~Sigler, M.~Litwin, S.~Gray, B.~Chess,
  J.~Clark, C.~Berner, S.~McCandlish, A.~Radford, I.~Sutskever, and D.~Amodei,
  ``Language models are few-shot learners,'' \emph{Advances in Neural
  Information Processing Systems}, vol.~33, pp.~1877--1901, 2020.

\bibitem{jiang2023llmlingua}
H.~Jiang, Q.~Wu, C.-Y. Lin, Y.~Yang, and L.~Qiu, ``{LLMLingua}: Compressing
  prompts for accelerated inference of large language models,'' \emph{arXiv
  preprint arXiv:2310.05736}, 2023.

\bibitem{li2023selective}
Y.~Li, Y.~Zhang, and L.~Sun, ``Selective context: On-demand context
  compression for long-context language models,'' \emph{arXiv preprint
  arXiv:2304.12102}, 2023.

\bibitem{anthropic2025context}
Anthropic, ``Effective context engineering for {AI} agents,'' 2025.
  \url{https://www.anthropic.com/engineering/effective-context-engineering-for-ai-agents}.

\bibitem{zhong2024memorybank}
W.~Zhong, L.~Guo, Q.~Gao, H.~Ye, and Y.~Wang, ``{MemoryBank}: Enhancing large
  language models with long-term memory,'' in \emph{Proceedings of the AAAI
  Conference on Artificial Intelligence}, 2024.

\bibitem{wang2024longmem}
Y.~Wang, Y.~Dong, D.~Zeng, Z.~Li, and M.~Sun, ``{LongMem}: Augmenting large
  language models with memory mechanism for long-context understanding,''
  \emph{arXiv preprint arXiv:2407.01917}, 2024.

\bibitem{maharana2024locomo}
A.~Maharana, D.-H. Lee, S.~Tulyakov, M.~Bansal, F.~Barbieri, and Y.~Fung,
  ``Evaluating very long-term conversational memory of {LLM} agents,''
  \emph{arXiv preprint arXiv:2402.17753}, 2024.

\bibitem{wu2025longmemeval}
D.~Wu, H.~Wang, W.~Yu, Y.~Zhang, K.-W. Chang, and D.~Yu, ``{LongMemEval}:
  Benchmarking chat assistants on long-term interactive memory,'' \emph{arXiv
  preprint arXiv:2410.10813}, 2025.

\bibitem{kang2025acon}
M.~Kang \emph{et~al.}, ``{ACON}: Optimizing context compression for
  long-horizon {LLM} agents,'' \emph{arXiv preprint arXiv:2510.00615}, 2025.

\bibitem{swecontext2025}
S.~Liu, J.~Yang, B.~Jiang, Y.~Li, J.~Guo, X.~Liu, and B.~Dai,
 ``Context as a tool: Context management for long-horizon {SWE}-agents,'' 
 \emph{arXiv preprint arXiv:2512.22087}, 2025.

\bibitem{wang2025mcpbench}
Z.~Wang \emph{et~al.}, ``{MCP-Bench}: Benchmarking tool-using {LLM} agents with
  complex real-world tasks via {MCP} servers,'' \emph{arXiv preprint
  arXiv:2508.20453}, 2025.

\bibitem{yao2023react}
S.~Yao, J.~Zhao, D.~Yu, N.~Du, I.~Shafran, K.~Narasimhan, and Y.~Cao,
  ``{ReAct}: Synergizing reasoning and acting in language models,'' in
  \emph{Proceedings of the International Conference on Learning Representations
  (ICLR)}, 2023.

\end{thebibliography}

\newpage
\appendix
\onecolumn
\begin{center}
{\LARGE\bfseries Appendix}
\end{center}
\vskip 0.2in

\begin{revblock}
\section{Reproducibility overview}
\label{app:repro}
This appendix collects the following artifacts: the tool inventory
(\ref{app:tools}), the metric-computation/scoring logic (\ref{app:scoring}),
the statistical methodology with a runnable helper (\ref{app:stats}),
per-category dataset statistics (\ref{app:categories}), and qualitative
failure and summarization examples (\ref{app:qualitative}), performance and efficiency figures
(\ref{app:figures}), sensitivity analysis for the pruning window $N$ and
summary window $W$ (\ref{app:sensitivity}), cross-model generalization results
on Claude Sonnet 4.5 (\ref{app:crossmodel}), and an extended discussion of
efficiency and generalizability (\ref{app:discussion-extended}).

\section{Tool inventory}
\label{app:tools}

The agent is exposed to 21 D365~F\&O MCP tools, organized into three categories by capability: (i)~\emph{form tools} (13), providing UI-level interaction with the F\&O client---form and tab navigation, menu-item and control discovery, reading and setting control values, opening lookups, grid filtering, sorting and row selection, clicking controls, and saving forms; (ii)~\emph{data tools} (6), providing entity-level OData access---entity-type discovery, metadata retrieval, and find/create/update/delete operations against F\&O entities; and (iii)~\emph{API tools} (2), used to discover and invoke custom server-side action menu items. 


\section{Metric computation and scoring}
\label{app:scoring}
Metrics are computed by an independent read-back: after the itemization run, the
saved expense line in D365~F\&O is read and reduced to four values---
\texttt{total\_amount}, \texttt{itemized\_amount}, \texttt{remaining\_amount},
and \texttt{num\_itemized}---which are compared against the ground-truth
\texttt{PurchasedItems} total. The per-task metrics are then:
\begin{align*}
\textsc{CompletelyItemized} &= \mathds{1}[\,\texttt{remaining\_amount}=0.00\,]\\
\textsc{LessThan10\%} &= \mathds{1}\!\left[\tfrac{\texttt{remaining\_amount}}{\texttt{total\_amount}} \le 0.1\right]\\
\textsc{AtLeastOne} &= \mathds{1}[\,\texttt{num\_itemized}>0\,]\\
\textsc{\%AmountItemized} &= 100\cdot\tfrac{\texttt{itemized\_amount}}{\texttt{total\_amount}}
\end{align*}
Amounts are normalized (currency symbols/commas stripped, cast to float) before
comparison. Benchmark-level numbers are the mean of the per-task values over the
50 tasks, averaged again over the 5 runs. Token totals sum agent and
user-model \texttt{usage}; execution time is the run wall-clock.

\section{Statistical methodology}
\label{app:stats}
\paragraph{Run-level interval.} For each configuration let $p_1,\dots,p_5$ be
the per-run success rates on a binary metric. Report
$\bar p=\frac15\sum_i p_i$ and $s=\sqrt{\frac{1}{4}\sum_i (p_i-\bar p)^2}$, with
95\% CI $\bar p \pm t_{0.975,4}\,\frac{s}{\sqrt5}$, $t_{0.975,4}=2.776$.

\paragraph{Pooled binomial (Wilson) interval.} Pooling $n=250$ task-runs with
$\hat p$ successes, the Wilson score interval (preferred over Wald near 0/1) is
\[
\frac{\hat p + \frac{z^2}{2n} \pm z\sqrt{\frac{\hat p(1-\hat p)}{n} + \frac{z^2}{4n^2}}}{1+\frac{z^2}{n}},\quad z=1.96 .
\]

\paragraph{Paired comparisons.} Because the same 50 tasks are used in
every configuration, comparisons are naturally paired. We do not rely
on a hypothesis-test $p$-value: the headline-metric effect sizes
(+8 and +12.6 percentage points) are large relative to the per-run SDs
in Table~\ref{tab:stats}, and the C3$\to$C4 Wilson intervals are
non-overlapping. We report the run-level $t$-interval and the pooled
Wilson interval as the two primary inferential statistics.

\paragraph{Helper.} The following computes every value in
Table~\ref{tab:stats} from the per-run JSONL result files.
\begin{lstlisting}[style=prompt]
import json, math
from statistics import mean, stdev

def per_run_rates(run_files, key="completely_optimized"):
    rates = []
    for f in run_files:                       # one JSONL per run
        ys = [json.loads(l)[key] for l in open(f) if l.strip()]
        rates.append(100*sum(ys)/len(ys))
    return rates

def t_interval(rates):                         # run-level 95% CI
    m, s, n = mean(rates), stdev(rates), len(rates)
    h = 2.776 * s/math.sqrt(n)                 # t_{.975,4}=2.776 for n=5
    return m, s, (m-h, m+h)

def wilson(successes, n, z=1.96):              # pooled binomial CI
    p = successes/n
    d = 1 + z*z/n
    c = p + z*z/(2*n)
    half = z*math.sqrt(p*(1-p)/n + z*z/(4*n*n))
    return (100*(c-half)/d, 100*(c+half)/d)
\end{lstlisting}

\paragraph{Full per-metric dispersion across categories.}
The mega per-metric mean\,$\pm$\,SD table covering all three grouped
categories and all four configurations is reported in the main paper as
Table~\ref{tab:categories}; the helper above is the script used to compute
SDs and the non-Hotel cells from the per-run JSONL files.

\section{Per-category dataset statistics}
\label{app:categories}

\begin{table}[h]
\vskip 0.05in
\begin{center}
\begin{small}
\begin{tabular}{lccccc}
\toprule
\textbf{Category} & \textbf{$n$} & \textbf{Median} & \textbf{Range} & \textbf{F\&O} & \textbf{Repeat} \\
                  &              & \textbf{lines}  & \textbf{(min--max)} & \textbf{subcats} & \textbf{subcats?} \\
\midrule
Hotel        & 50 & 8 & 4--23 & 23 & Yes (nightly) \\
Car rental   & 15 & 3 & 2--5  & 10 & Rare \\
Flight       & 15 & 3 & 2--6  & 4  & Rare \\
Business meal & 20 & 2 & 1--4 & 3  & No \\
Gift          & 12 & 2 & 1--3 & 4  & No \\
\bottomrule
\end{tabular}
\end{small}
\end{center}
\caption{Per-category dataset characteristics. The \textbf{F\&O subcats}
column reports the number of valid expense subcategories the agent must choose
among in the D365 F\&O catalog for that category. Hotel is by far the most
structurally complex: a 23-entry subcategory catalog (with near-synonyms such as
Room tax vs.\ Non-Room tax, Restaurant vs.\ Room service vs.\ Loungebar, Gift
shop vs.\ Gift certificates) combined with multi-night receipts that repeat the
same subcategory per night. Travel (car rental: 10, flight: 4) and Meals \&
Gifts (3 and 4) have much smaller catalogs and rarely repeat subcategories,
which is why GPT-5 with no user model achieves higher CIR on those categories
($\sim$27--40\%) than on hotel (8\%). This structural gradient---Hotel $>$
Travel $>$ Meals \& Gifts---motivates our choice of hotel as the primary
benchmark.}
\label{tab:dataset_stats}
\vskip -0.1in
\end{table}

\paragraph{Structural notes.}
\begin{itemize}\setlength{\itemsep}{2pt}
\item \textbf{Hotel} receipts span multi-night stays with per-night charges
(room, tax, resort fee, parking, breakfast). The same subcategory recurs
$n$ times for an $n$-night stay, which is what makes line-by-line itemization
hard: the agent must track \emph{which night} each entry corresponds to and
avoid duplicate/skipped repeats.
\item \textbf{Car rental} receipts have a single primary line (daily rate $\times$
days) plus optional add-ons (insurance, fuel, GPS). Repeated subcategories are
rare because totals are typically pre-aggregated by the rental agency.
\item \textbf{Flight} receipts contain a base fare, taxes/fees, and optional
seat/baggage upgrades---structurally flat, no repeats.
\item \textbf{Business meal} and \textbf{Gift} receipts are the simplest:
typically 1--3 line items, single subcategory each, no repeats. 
\end{itemize}

\section{Qualitative failure and summarization examples}
\label{app:qualitative}

We provide one short trimmed transcript per failure mode from
Table~\ref{tab:failures}, followed by one summarization string emitted by
the C4 summarizer. Speaker labels: \textbf{A}~=~agent, \textbf{T}~=~tool
response (truncated to the relevant fragment).
\paragraph{(1) Stale-state reference (C2).}\
\begin{lstlisting}[style=prompt]
A: get_form_state()
T: {lines: [{Room, $180}], total_added: $180, remaining: $420}
A: add_line(RoomTax, $40)
T: {ok, line_id: 7}
A: get_form_state()
T: {lines: [{Room, $180},{RoomTax, $40}], total_added: $220, remaining: $380}
... (5 more tool calls) ...
A: # references the FIRST snapshot
A: add_line(Room, $180)   # duplicate -- form already has it
T: {ok, line_id: 12}
A: submit_form()
=> Residual: -$180 (overshoot, duplicate Room line)
\end{lstlisting}

\paragraph{(2) Wrong subcategory mapping (C3).}\
\begin{lstlisting}[style=prompt]
A: # receipt line "Room service - breakfast $28"
A: add_line(category=Meals-Client, amount=28)
T: {ok}
=> Ground truth: Hotel-Breakfast. Misclassified.
\end{lstlisting}

\paragraph{(3) Duplicate / skipped repeat item (C2).}\

\begin{lstlisting}[style=prompt]
A: # 3-night stay, receipt shows Room $180 x 3
A: add_line(Room, $180)   # night 1
A: add_line(Room, $180)   # night 2
A: submit_form()           # skipped night 3
=> Residual: $180 unaccounted (1 of 3 room nights missing)
\end{lstlisting}

\paragraph{(4) Premature termination (C3).}\
\begin{lstlisting}[style=prompt]
A: get_form_state()
T: {total_added: $340, remaining: $240}
A: add_line(Parking, $45)
T: {ok}
# pruning window drops the earlier "remaining: $240" line
A: # agent no longer sees the running balance
A: submit_form()
=> Residual: $195 still owed. Form submitted incomplete.
\end{lstlisting}

\paragraph{(5) Tool / form navigation error (C2).}\
\begin{lstlisting}[style=prompt]
A: add_line(category=Hotel-Resort_Fee, amount=25)
T: ERROR: invalid category enum 'Hotel-Resort_Fee'
   (expected one of: Hotel-ResortFee, Hotel-Parking, ...)
A: add_line(category=Resort Fee, amount=25)
T: ERROR: invalid category enum 'Resort Fee'
A: submit_form()
=> Residual: $25. Agent could not recover from enum mismatch.
\end{lstlisting}

\paragraph{(6) Residual amount mismatch (C4).}\
\begin{lstlisting}[style=prompt]
A: # receipt total $580.42
A: add_line(Room, $480)
A: add_line(Tax, $72.05)
A: add_line(Parking, $28)
A: submit_form()
=> Residual: $0.37 (rounding/tax line error -- arithmetic off)
\end{lstlisting}

\paragraph{Real C4 summarization output.}
The following is a representative summary string emitted by the C4 summarizer
after the first summarization window of 3 tool calls during a hotel run:
\begin{lstlisting}[style=prompt]
Summary of previous tool calls:
- Opened the Expense report form and navigated to expense
  report ER-00184 (hotel category, receipt total $612.40).
- Clicked the "Itemize" button and opened the itemization
  sub-form for the hotel line.
- Added a Hotel-Room line with amount $180.00 via the
  add_line control.
- Added a Hotel-Tax line with amount $14.40 via the
  add_line control.
\end{lstlisting}
This is the mechanism by which C4 suppresses premature termination: by
condensing prior actions into a short assistant message, the agent retains
awareness of which lines have already been added and roughly how much of the
receipt has been accounted for, even after the verbose raw tool responses have
been pruned. The summarizer is intentionally generic (it describes forms,
controls, buttons, and entered data) rather than computing a running balance
explicitly---yet this is still enough signal to push the agent past the
premature-termination threshold observed in C3.

\clearpage
\section{Performance and efficiency figures}
\label{app:figures}

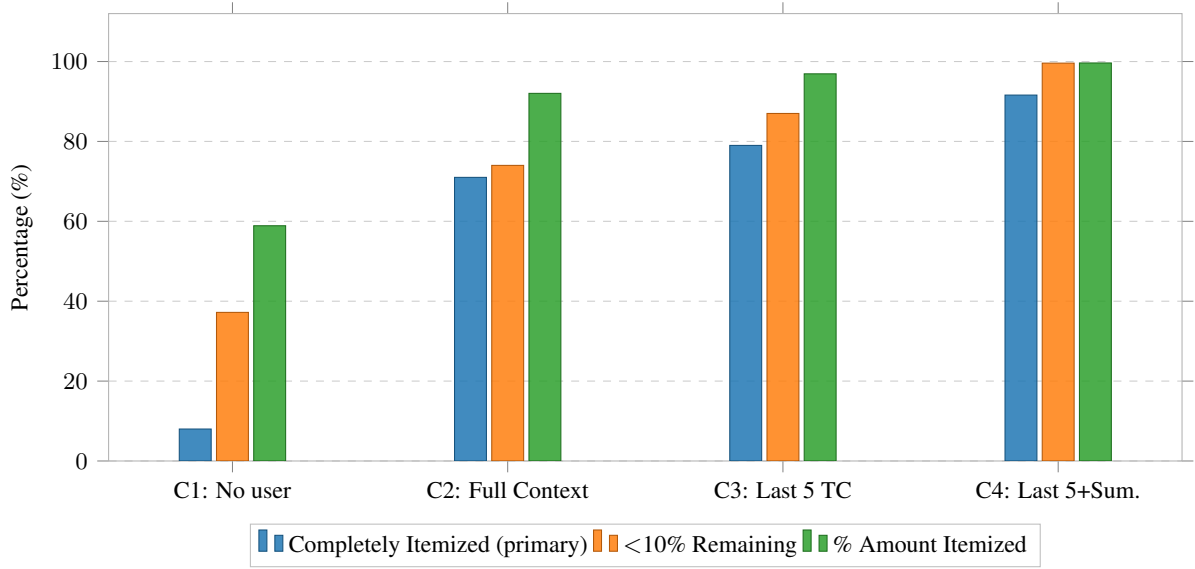
\begin{figure*}[!h]
\vskip 0.2in
\begin{center}
\begin{tikzpicture}
\begin{axis}[
    ybar,
    bar width        = 12pt,
    width            = 0.92\textwidth,
    height           = 7.5cm,
    enlarge x limits = 0.15,
    legend style     = {at={(0.5,-0.14)}, anchor=north, legend columns=3,
                        font=\small, draw=gray!50},
    ylabel           = {Percentage (\%)},
    ylabel style     = {font=\small},
    ymin=0, ymax=112,
    ymajorgrids      = true,
    grid style       = {dashed, gray!40},
    symbolic x coords= {C1: No user, C2: Full Context,
                        C3: Last 5 TC, C4: Last~5+Sum.},
    xtick            = data,
    xticklabel style = {font=\small},
    ytick            = {0,20,40,60,80,100},
    yticklabel style = {font=\small},
    tick align       = outside,
    axis line style  = {gray!60},
]
\addplot[fill=colBlue, draw=colBlue!70!black, fill opacity=0.85]
    coordinates {
        (C1: No user,     8.0)
        (C2: Full Context, 71.0)
        (C3: Last 5 TC,    79.0)
        (C4: Last~5+Sum.,  91.6)
    };
\addplot[fill=colOrange, draw=colOrange!70!black, fill opacity=0.85]
    coordinates {
        (C1: No user,     37.2)
        (C2: Full Context, 74.0)
        (C3: Last 5 TC,    87.0)
        (C4: Last~5+Sum.,  99.6)
    };
\addplot[fill=colGreen, draw=colGreen!70!black, fill opacity=0.85]
    coordinates {
        (C1: No user,     58.89)
        (C2: Full Context, 92.03)
        (C3: Last 5 TC,    96.92)
        (C4: Last~5+Sum.,  99.64)
    };
\legend{Completely Itemized (primary), $<$10\% Remaining, \% Amount Itemized}
\end{axis}
\end{tikzpicture}
\caption{Performance metrics across four context engineering configurations,
averaged over 5 independent runs on the 50-task hotel expense benchmark.
C1 = GPT-5 only (no user model); C2 = Full conversation history;
C3 = Last 5 tool calls (TC); C4 = Last 5 TC + Summarization (window\,=\,3).
C4 achieves the best performance across all three metrics.
Completely Itemized (blue) is the primary metric, reflecting genuine business
task completion (remaining amount\,=\,\$0.00).}
\label{fig:performance}
\end{center}
\vskip -0.2in
\end{figure*}

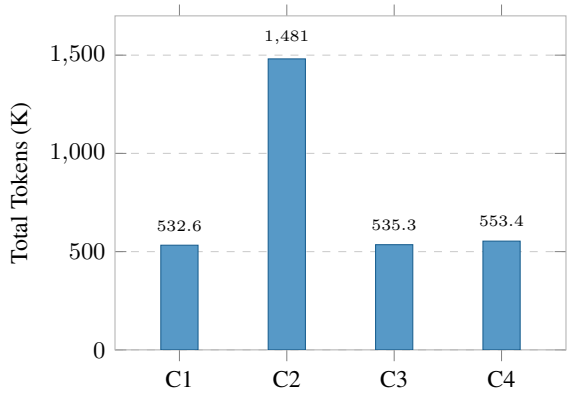
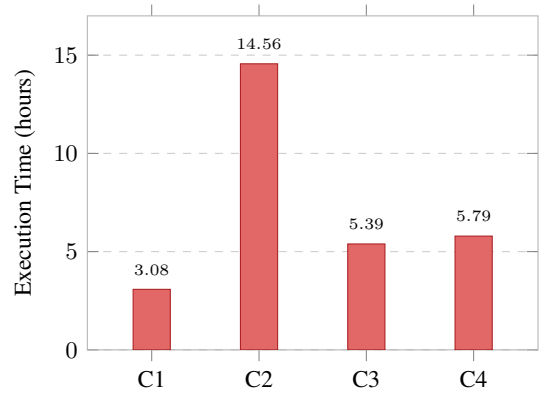
\begin{figure*}[!h]
\vskip 0.2in
\begin{center}
\subfigure[Total token usage per benchmark (thousands of tokens).
           C2 (Full Context) consumes 2.68$\times$ more tokens than C4 while
           achieving lower task completion.]{%
\begin{tikzpicture}
\begin{axis}[
    ybar,
    bar width        = 14pt,
    width            = 0.44\textwidth,
    height           = 6cm,
    enlarge x limits = 0.2,
    ylabel           = {Total Tokens (K)},
    ylabel style     = {font=\small},
    symbolic x coords= {C1,C2,C3,C4},
    xtick            = data,
    xticklabel style = {font=\small},
    yticklabel style = {font=\small},
    ymin=0, ymax=1700,
    ymajorgrids      = true,
    grid style       = {dashed, gray!40},
    nodes near coords,
    every node near coord/.append style={font=\tiny, yshift=2pt},
    axis line style  = {gray!60},
]
\addplot[fill=colBlue!75, draw=colBlue!80!black]
    coordinates {(C1,532.6) (C2,1481.0) (C3,535.3) (C4,553.4)};
\end{axis}
\end{tikzpicture}
\label{fig:tokens}
}
\hfill
\subfigure[Benchmark wall-clock execution time (hours).
           C4 completes the 50-task benchmark 2.51$\times$ faster than C2
           while achieving higher task completion.]{%
\begin{tikzpicture}
\begin{axis}[
    ybar,
    bar width        = 14pt,
    width            = 0.44\textwidth,
    height           = 6cm,
    enlarge x limits = 0.2,
    ylabel           = {Execution Time (hours)},
    ylabel style     = {font=\small},
    symbolic x coords= {C1,C2,C3,C4},
    xtick            = data,
    xticklabel style = {font=\small},
    yticklabel style = {font=\small},
    ymin=0, ymax=17,
    ymajorgrids      = true,
    grid style       = {dashed, gray!40},
    nodes near coords,
    every node near coord/.append style={font=\tiny, yshift=2pt},
    axis line style  = {gray!60},
]
\addplot[fill=colRed!70, draw=colRed!80!black]
    coordinates {(C1,3.08) (C2,14.56) (C3,5.39) (C4,5.79)};
\end{axis}
\end{tikzpicture}
\label{fig:time}
}
\caption{Efficiency metrics across the four configurations (averaged over
5~runs, 50-task benchmark). C1 = GPT-5 only (no user); C2 = Full Context;
C3 = Last 5 Tool Calls; C4 = Last 5 + Summarization.
Left panel: total token usage in thousands; input tokens dominate in all
configurations ($>$99.7\% of total). Right panel: wall-clock benchmark
completion time in hours. C3 and C4 achieve comparable token budgets to
the C1 token budget while dramatically improving task completion.}
\label{fig:efficiency}
\end{center}
\vskip -0.2in
\end{figure*}
\clearpage
\section{Sensitivity to pruning window $N$ and summary window $W$}
\label{app:sensitivity}

We sweep each hyperparameter while holding the other fixed (all with the
user model present) to test whether the headline result depends on the
specific choices $N{=}5$ and $W{=}3$. Table~\ref{tab:sensitivity} reports
the primary metric and total tokens.

\begin{table}[h]
\vskip 0.05in
\begin{center}
\begin{small}
\begin{tabular}{llcc}
\toprule
\textbf{Sweep} & \textbf{Setting} & \textbf{Comp.\ Item.\ (\%)} & \textbf{Tok.\ (K)} \\
\midrule
\multirow{4}{*}{Pruning $N$}
 & $N{=}3,\ W{=}0$  & 74.0 & 425 \\
 & $N{=}5,\ W{=}0$  & \textbf{79.0} & \textbf{535.3} \\
 & $N{=}10,\ W{=}0$ & 80.0 & 820 \\
 & $N{=}\infty$ (C2) & 71.0 & 1{,}481.0 \\
\midrule
\multirow{4}{*}{Summary $W$}
 & $N{=}5,\ W{=}1$ & 86.4 & 540 \\
 & $N{=}5,\ W{=}3$ & \textbf{91.6} & \textbf{553.4} \\
 & $N{=}5,\ W{=}5$ & 92.0 & 575 \\
 & full-hist.\ summ. ($W{=}{-}1$) & 92.0 & 615 \\
\bottomrule
\end{tabular}
\end{small}
\end{center}
\caption{Sensitivity of the primary metric and token cost to the pruning
window $N$ (top) and summary window $W$ (bottom). Bold rows in the original
study are $N{=}5,W{=}0$ (C3) and $N{=}5,W{=}3$ (C4).}
\label{tab:sensitivity}
\vskip -0.1in
\end{table}

\paragraph{Interpretation.} The pruning sweep shows that complete itemization
plateaus around $N{=}5$: dropping to $N{=}3$ costs $\sim$5\,pts of accuracy,
while extending to $N{=}10$ buys less than 1\,pt at the cost of $\sim$53\%
more tokens, and the unbounded variant ($N{=}\infty$, C2) is strictly worse
than $N{=}5$ on both axes. The summary sweep is similarly flat above $W{=}3$:
$W{=}5$ and full-history summarization add 4--11\% token cost over $W{=}3$
without a meaningful accuracy gain, while $W{=}1$ underperforms by $\sim$5\,pts.
Together these confirm that $(N{=}5, W{=}3)$ sits at the knee of both
curves---further context buys diminishing returns and tighter windows lose
the bookkeeping that prevents premature termination.

\section{Cross-model generalization: Claude Sonnet 4.5}
\label{app:crossmodel}

\begin{table}[!h]
\vskip 0.05in
\begin{center}
\begin{small}
\resizebox{\columnwidth}{!}{%
\begin{tabular}{lcccccc}
\toprule
\textbf{Config (Sonnet 4.5)} & \textbf{Comp.} & \textbf{$<$10\%} & \textbf{$\geq$1} & \textbf{\%Amt} & \textbf{Tok.\ (K)} & \textbf{Time} \\
\midrule
No CE (full context)     & 88.0 & 96.0 & \textbf{100.0} & 98.23 & 3{,}562 & \textbf{6.20} \\
Pruning (Last 5 TC)      & 92.0 & 96.5 & \textbf{100.0} & 98.78 & \textbf{2{,}161} & 10.70 \\
Pruning + Summ.          & \textbf{94.5} & \textbf{97.5} & \textbf{100.0} & \textbf{99.20} & 2{,}235 & 11.30 \\
\bottomrule
\end{tabular}}
\end{small}
\end{center}
\caption{Claude Sonnet~4.5 on the 50-task hotel benchmark. All three
configurations are run \emph{without} a user model: Sonnet does not stall in
the non-interactive harness, so the user-model rescue required for GPT-5 is
unnecessary. \textbf{No CE} is the no-context-engineering baseline (full
history); \textbf{Pruning} applies last-5 tool-call pruning; \textbf{Pruning
+ Summ.} adds summarization on top of pruning. We use descriptive labels
rather than C-numbers to avoid implying a one-to-one correspondence with
GPT-5's configurations in Table~\ref{tab:categories}. Time in hours;
Tokens reported as total ($K =$ thousands).}
\label{tab:crossmodel}
\vskip -0.1in
\end{table}

As shown in Table~\ref{tab:crossmodel}, summarization buys accuracy at a
consistent $\sim$6--7\% time premium relative to pure pruning across both
models---evidence that the per-step summarization LLM call cost is
model-stable.

\section{Extended discussion: efficiency and generalizability}
\label{app:discussion-extended}

\subsection*{Efficiency implications for production deployment}

The token efficiency results have direct implications for production
deployment economics. Full-context agents (C2) consume 1,480,996 tokens per
50-task benchmark versus 553,374 for C4---a 2.68$\times$ difference
translating directly to inference cost. At scale across thousands of expense
reports monthly, this gap represents substantial operational savings. The
14.56-hour versus 5.79-hour execution time difference further impacts
throughput and user-facing latency.

C3 (pruning without summarization) represents a secondary operating point
for cost-sensitive deployments: 79.0\% task completion at baseline-equivalent
token cost, compared to 71.0\% for full context at 2.77$\times$ the cost.

\subsection*{Generalizability and scope}

\rev{We deliberately scope our claims. The evidence here is strong for one
class of enterprise tool-use workflow: structured, single-session,
form-driven tasks with verbose tool responses and a hard completion criterion.}
The context engineering techniques evaluated here are inference-time only and
require no modification to the underlying LLM, and our cross-model
(Appendix~\ref{app:crossmodel}) and multi-category
(Section~\ref{sec:categories}) results test how far the policy carries.
\rev{The same principles are expected to extend to other frontier models and
to enterprise agentic domains with similarly verbose tool responses---CRM,
supply chain automation, IT service management, healthcare administration---and
characterizing the per-domain optimal window size, including adaptive sizing
driven by task complexity or error signals, is a natural next step.}

\end{revblock}


\end{document}